\documentclass[journal]{IEEEtran}
\ifCLASSINFOpdf
\else
\fi
\ifCLASSINFOpdf
\usepackage[pdftex]{graphicx}
\else
\fi

\usepackage{amsmath}

\usepackage{array}
\usepackage{url}
\usepackage{graphicx}
\usepackage{amsmath}
\usepackage{amssymb}

\usepackage{fmtcount}
\usepackage{color}
\usepackage{multicol}
\usepackage{multirow} 
\usepackage{algorithmic}
\usepackage[ruled,lined,linesnumbered]{algorithm2e}

\begin{document}
%
\title{Semantic and Contrast-Aware Saliency}
%
%
%

\author{Xiaoshuai~Sun
\thanks{This work was supported in part by the National Natural Science Foundation of China under Project 61472103, 61572108 and 61632007, and in part by the Australian Research Council under Grant FT130101530.
	
Xiaoshuai Sun was with the School of Computer Science and Technology, Harbin Institute of Technology, Harbin, China, email: \{xiaoshuaisun\}@hit.edu.cn.
}
}

\maketitle

\begin{abstract}
In this paper, we proposed an integrated model of semantic-aware and contrast-aware saliency combining both bottom-up and top-down cues for effective saliency estimation and eye fixation prediction. The proposed model processes visual information using two pathways. The first pathway aims to capture the attractive semantic information in images, especially for the presence of meaningful objects and object parts such as human faces. The second pathway is based on multi-scale on-line feature learning and information maximization, which learns an adaptive sparse representation for the input and discovers the high contrast salient patterns within the image context. The two pathways characterize both long-term and short-term attention cues and are integrated dynamically using maxima normalization. We investigate two different implementations of the semantic pathway including an End-to-End deep neural network solution and a dynamic feature integration solution, resulting in the SCA and SCAFI model respectively. Experimental results on artificial images and 5 popular benchmark datasets demonstrate the superior performance and better plausibility of the proposed model over both classic approaches and recent deep models.

\end{abstract}

\begin{IEEEkeywords}
Saliency, feature integration, eye fixation prediction.
\end{IEEEkeywords}

%
\IEEEpeerreviewmaketitle

\section{Introduction}
\label{sec:intro}
\IEEEPARstart{T}{he} last two decades have witnessed enormous development in the field of computational visual attention modeling and saliency detection \cite{borji2013state}. Various models, datasets, and evaluation metrics are proposed to help machines better understand and predict human viewing behavior. By producing a 2D/3D saliency map that predicts where human look, a computational model can be applied to many low-level computer vision applications, e.g. image retargeting \cite{goferman2012context}, video compression  \cite{guo2010novel,hadizadeh2014saliency}, detecting abnormal patterns \cite{Itti_etal98pami}, segmenting proto objects \cite{Hou2007saliencydetection}, generating object proposals  \cite{alexe2012measuring}.

\begin{figure}[t]
	\begin{center}
		\includegraphics[width=\linewidth]{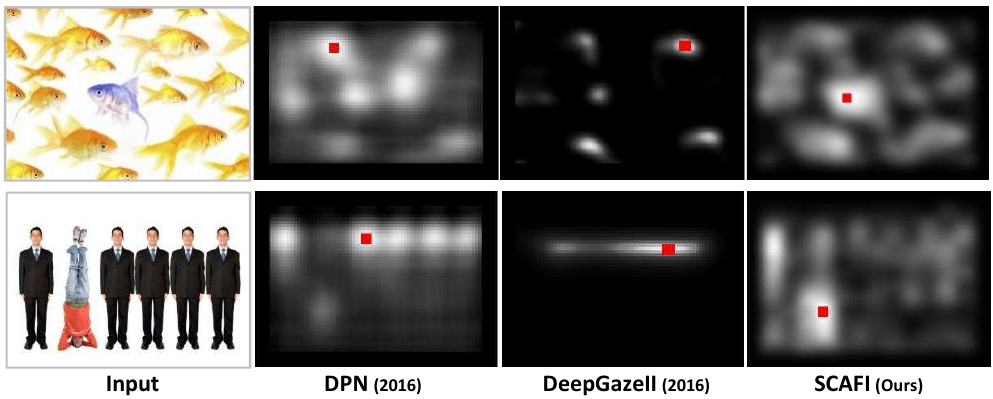}
		\vspace{-0.5cm}
	\end{center}
	\caption{Illustration of our model. For each row, we show the input image (left), the saliency prediction result of \textbf{DPN} \cite{pan2016shallow}, \textbf{DeepGazeII} \cite{kummerer2016deepgaze} and our proposed \textbf{SCAFI} model. The red dot in each saliency map indicates the location of the maximum value. Compared to traditional deep models, our method highlights both the semantic-aware and contrast-aware saliency.}
	\label{fig:Motivation}
\end{figure}

The concept of ``saliency'' were investigated not only in early vision modeling but also in many engineering applications such as image compression  \cite{itti2004automatic}, object recognition \cite{salah2002selective} and tracking \cite{frintrop2010general}, robot navigation \cite{siagian2009biologically}, design and advertising \cite{rosenholtz2011predictions} etc. 
In the early stage, the research works of attention and saliency are mostly motivated by biological priors  \cite{Itti_etal98pami,koch1987shifts}, statistical assumptions \cite{gao2008plausibility}， frequency domain characteristics \cite{Hou2007saliencydetection}, \cite{hou2012image} and information theory \cite{Bruce_Tsotsos06nips,Hou08nips}. Some papers also draw their inspirations by analyzing the statistical property of ground-truth eye-tracking data \cite{sun2012saliency,sun2014toward,garcia2012relationship}. 

Driven by the great success of deep neural networks, the deep learning scheme were recently introduced to saliency research and undoubtfully achieved ground-break success in predicting human eye-fixations \cite{kummerer2014deep,huang2015salicon,pan2016shallow}. The remarkable performance of deep models relies very much on their ability of characterizing semantic information in the inputs. However, they can hardly explain the psychophysical evidence, and often produce unreasonable responses to obvious salient patterns. For instance, in Figure \ref{fig:Motivation}, the saliency maps produced by \textbf{DPN} \cite{pan2016shallow} and DeepGazeII \cite{kummerer2016deepgaze} only characterize the presence of meaningful objects, e.g. faces and fish, but fail to localize the most salient pattern in the image.

As discussed in \cite{bruce2016deeper}, some eye fixations are directed at objects, while others are attracted by local feature contrast which is relatively detached from semantics. It's crucial to consider both semantic and contrast cues for the construction of a plausible and effective saliency model. On one hand, the integration of deep neural network is inevitable, since traditional methods can not discover or make effective use of large-scale features that represent attractive semantic objects or common interests. On the other hand, we must carefully formulate the model to ensure it is truely a model of saliency instead of a simplified object detector derived from recognition models. Based on the above considerations, we have proposed an integrated saliency model, namely \textbf{SCA} \cite{sun2017saliency}, which combines both semantic and contrast cues for effective saliency estimation and eye-fixation prediction. \textbf{SCA} contains two pathways and responses to both bottom-up and top-down cues. In the experiments, we demonstrated that our \textbf{SCA} model steadily outperforms both the traditional saliency models and the cutting-edge End-To-End deep models. Despite its superior performance, \textbf{SCA} still has its own limitations: 

1) It only inherits the initial convolutional layers of the \textbf{VGG} model which might cause \textit{semantic loss} during the generation of semantic-aware saliency; 

2) It requires large-scale eye-fixation data for the training of the semantic pathway.

In this paper, to overcome the limitations of \textbf{SCA}, we proposed a new model called \textbf{SCAFI} which predicts saliency based on semantic and contrast-aware feature integration. \textbf{SCAFI} is a heuristic model which shares the same inspiration of \textbf{SCA}, but achieves much better performance by directly pooling the semantic information out of the \textbf{VGG} net. Compared to traditional deep model (Figure \ref{fig:Motivation}), \textbf{SCAFI} is learning-free and highlights both the semantic interests and the bottom-up contrast. The main contributions of our work are listed as follows:

\begin{itemize}
	\item{We propose an heuristic saliency model (\textbf{SCAFI} ) that characterizes both top-down and bottom-up saliency cues in a unified framework.}
	\item{Compared to our previous model \textbf{SCA}, \textbf{SCAFI} directly extracts the semantic information from the \textbf{VGG} net, which avoids the fine-tuning process and thus does not require additional labeled data to perform the eye-fixation prediction task.}
	\item{The \textbf{SCAFI} model outperforms the state-of-art methods, including recent deep models (e.g. \textbf{DPN}, winner of LSUN Challenge 2015) and our previous model \textbf{SCA}, on all 5 eye-fixation datasets.}
	\item{Despite its superior performance in eye fixation prediction task, the \textbf{SCAFI} model also produces plausible responses to images with pre-attentive patterns and high-contrast objects.}
\end{itemize}

The following parts of the paper are organized as follows. Sec. \ref{sec:rw} introduces the most related works in detail. Sec. \ref{sec:tm} presents our \textbf{SCA} and \textbf{SCAFI} model and some intermediate results. In Sec.\ref{sec:exp}, we show the main experimental results on several well-acknowledged datasets and provided more comparisons with different parameter settings. Finally, we conclude our work in Sec.\ref{sec:con}.

\section{Related Works}
\label{sec:rw}
There are many excellent research works contributed to this topic, most of them can be found in recent surveys \cite{borji2013state,li2014secrets}. Saliency models can be classified into two categories: ones driven by the task of \textit{salient object segmentation}, and the others driven by the task of \textit{human eye-fixation prediction}. 
The two tasks have their own benchmark datasets and preferred evaluation metrics. Our paper focuses on the fixation prediction task, thus we didn’t compared to the segmentation methods such as \textbf{RFCNs} \cite{wang2016saliency}, \textbf{DHSNet} \cite{liu2016dhsnet}, \textbf{DISC} \cite{chen2016disc}, and didn’t use PR-curves and F-measure metrics for the evaluation. Instead, we follow the work of Bruce \textit{et al.} \cite{bruce2016deeper} and Lebor{\'a}n \textit{et al}. \cite{leboran2017dynamic}, and use the fixation datasets and shuffled \textbf{AUC} metric for the evaluation. In this section, we mainly introduce the related works that focus on eye-fixation prediction task including the most related traditional models, along with some recently proposed deep-learning models. 

\subsection{Traditional Heuristic Saliency Model}
Itti \textit{et al.} \cite{Itti_etal98pami} implemented the very first computational model (\textbf{ITTI}) that generates 2D saliency map based on the Feature Integration theory \cite{treisman1980feature}. The center-surround operation inspires many subsequent models \cite{gao2008plausibility}. Graph-based Visual Saliency (\textbf{GBVS}) model \cite{harel2006graph} adopts the same bio-inspired features in \cite{Itti_etal98pami} yet a different parallelized graph computation measurement to compute visual saliency.   

Based on sparse representation, Bruce and Tsotsos \cite{Bruce_Tsotsos06nips} proposed the \textbf{AIM} model (Attention by Information Maximization) which adopts the self-information of ICA coefficients as the measure for signal saliency. Also based on information theory and sparse coding, Hou and Zhang \cite{Hou08nips} proposed the dynamic visual attention (\textbf{DVA}) model which defines spatio-temporal saliency as incremental coding length. Garcia \textit{et al. }\cite{garcia2012relationship,leboran2017dynamic} proposed the \textbf{AWS}(Adaptive Whitening Saliency) model which relies on a contextually adapted representation produced through adaptive whitening of color and scale features. Inspired by the success of online sparse representation, we also present a unified statistical framework named \textbf{SGP} \cite{sun2012saliency,sun2014toward} (Super-Gaussian Component Pursuit) which exploits statistical priors for the joint modeling of both saccadic eye movements and visual saliency.

Hou \textit{et al.} \cite{Hou2007saliencydetection} proposed a highly efficient saliency detection algorithm by exploring the spectral residual (\textbf{SR})  in the frequency domain. \textbf{SR} highlights the salient regions by manipulating the amplitude spectrum of the images' Fourier transformation. Inspired by \cite{Hou2007saliencydetection}, Guo \emph{et al.} \cite{Guo2008Spatio} achieved fast and robust spatio-temporal saliency detection by using the phase spectrum of Quaternion Fourier Transform. As a theoretical revision of \textbf{SR}, \cite{hou2012image} proposed a new visual descriptor named Image Signature (\textbf{SIG}) based on the Discrete Fourier Transform of images, which was proved to be more effective in explaining saccadic eye movements and change blindness of human vision.

Despite the above models, there are many other insightful works that detect visual saliency using different types of measures, e.g. Bayesian Surprise  \cite{itti2009bayesian}, Center-Surround Discriminant Power \cite{gao2008discriminant}, Short-Term Self-Information \cite{sun2010saliency}, Spatially Weighted Dissimilarity \cite{duan2011visual}, Site Entropy Rate  \cite{wang2010measuring}, Rarity \cite{riche2013rare2012} and Self-Resemblance \cite{seo2009static}. 
 
\subsection{Learning-based Saliency Model via Deep Neural Network}  
Provided with enough training data, deep model can achieve ground-breaking performance that are far better than traditional methods, sometime even outperform humans. The ensembles of Deep Networks (\textbf{eDN}) \cite{vig2014large} is the first attemp at modeling saiency with deep models, which combines three different convnet layers using a linear classifer. Different from \textbf{eDN}, recent models such as \textbf{DeepGaze}  \cite{kummerer2014deep}, \textbf{SALICON} \cite{huang2015salicon} and \textbf{FOCUS} \cite{bruce2016deeper} integrate pre-trained layers from large-scale CNN models. Especially, \textbf{SALICON} \cite{huang2015salicon}, \textbf{DeepFix} \cite{kruthiventi2015deepfix}  and  \textbf{DPN} \cite{pan2016shallow} use the \textbf{VGG} network pre-trained on ImageNet to initialize their convolutional layers and then train the rest layers based on ground-truth saliency maps generated using human eye-fixation data. 

Benefit from the powerful visual representation embedded in \textbf{VGG} net, the above models significantly outperform traditional methods in eye-fixation prediction task on almost all benchmark datasets. In this paper, we mainly compared our model with \textbf{SALICON}\cite{huang2015salicon}, \textbf{DeepGazeII}\cite{kummerer2016deepgaze} and \textbf{DPN}\cite{pan2016shallow}. \textbf{SALICON}\footnote{\textbf{SALICON}: \url{http://salicon.net/demo/#}}  and \textbf{DeepGazeII}\footnote{\textbf{DeepGazeII}: \url{https://deepgaze.bethgelab.org/}} provide on-line Web service to receive image submissions and generate saliency maps. \textbf{DPN}\footnote{\textbf{DPN}: \url{https://github.com/imatge-upc/saliency-2016-cvpr}} is a fully open source End-to-End model with good computation efficiency and state-of-the-art performance.

\subsection{Relationship between Previous Methods and Ours}
The success of the above-mentioned deep models is undoubtfully remarkable. The ground-breaking improvements achieved by integrating the pre-trained convolutional layers directly motivates our \textbf{SCA} model. However, the fine-tuning of the deep network is not an easy task, and most of the current models can not reliably detect and pop out high-contrast signal in the input context, even with a large-scale training set such as \textbf{SALICON} \cite{jiang2015salicon}. In contrast, traditional adaptive methods such as \textbf{AWS} \cite{garcia2012relationship} and \textbf{SGP} \cite{sun2014toward} go through a pure bottom-up pathway and are hardly affected by the dense presence of semantics. This further inspires us to formulate our \textbf{SCAFI} model which handles top-down semantic cues and the bottom-up contrast in two independent pathways.

\section{The Proposed Model}
\label{sec:tm}
In this section, we first formulate our problem and then introduce our integrated saliency model \textbf{SCAFI} along with its alternative version \textbf{SCA}. We also present the implementation details and the intermediate results of the proposed models. 

\begin{figure*}[t]
	\begin{center}
		\includegraphics[width=0.92\linewidth]{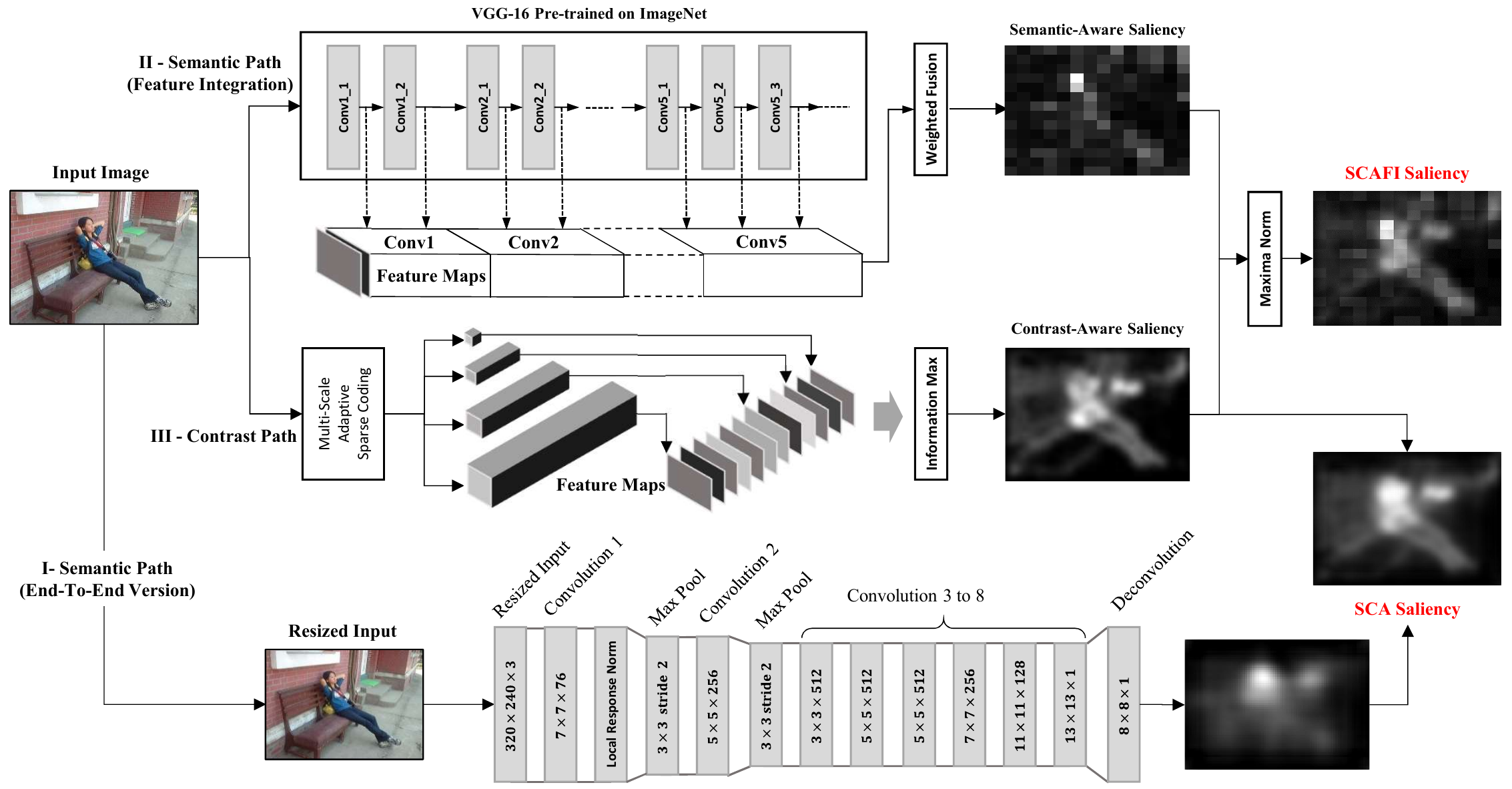}
	\end{center}
	\caption{Semantic and Contrast-Aware Saliency Model. \textbf{Part I }: An End-to-End deep neural network customized as an alternative version of the semantic-aware saliency (\textbf{SAS}). \textbf{Part II:} A deep feature integration module as the default \textbf{SAS} pathway, which directly collects and aggregates 2D feature maps from the convolutional layers of pre-trained VGG net \cite{chatfield2014return}. \textbf{Part III}: A contrast-aware saliency (\textbf{CAS}) pathway that discovers high contrast salient patterns within the image context based on multi-scale sparse representation and information maximization.   The temporal outputs of the \textbf{SAS} and \textbf{CAS} pathways are all resized to the original image size and then integrated using maxima normalization. The integration of \textbf{I} and \textbf{III }corresponds to our previous model \textbf{SCA}, while the combination of \textbf{II} and \textbf{III} results in our new model \textbf{SCAFI}. The experiments show that the new \textbf{SCAFI} model performs much better than the \textbf{SCA} model as well as the other state-of-the-arts. }
	\label{fig:Overview}
\end{figure*}

\subsection{Problem Defination}
Let $I$ be an image, $Y_{eye}$ be a set of pixel locations recording human eye fixations, and $Y_{rnd}$ a set of random or pseudo-random pixel locations. The task of eye fixation prediction is to generate a saliency map $S$ which predicts the probability of every pixel in the image for being a true human eye fixation. Eye-fixation prediction is initially proposed for the evaluation of computational visual attention models. Since visual attention can be driven by both bottom-up and top-down cues \cite{borji2013state}, saliency detection methods can also be divided into two main categories: 1. Unsupervised methods constructed based on biological and psychological knowledge of the human vision system; 2. Supervised methods which adopt machine learning techniques to train a saliency model based on visual datasets with eye tracking labels. 

\subsubsection{Saliency Detection as a Bottom-up Problem}
Salient stimuli often have the capability of attracting visual attention toward their locations in a pure bottom-up manner. Such kind of saliency is a natural property of the presented stimuli, which is independent of specific tasks or goals and can be detected by a fixed computational model. Bottom-up saliency detection approaches attract a lot of attention between the 1980s and 2010s. On one hand, they are theoretically trustful since most of them are the computational implementation of some well-studied psychological theories and hypotheses, e.g. Feature Integration \cite{treisman1980feature} and Information Maximization \cite{Bruce_Tsotsos06nips}. On the other hand, these models indeed generate reasonable responses to various controlled stimuli compared with those from human subjects. However, bottom-up methods can hardly model the effect of semantic factors in natural scenes such as the presence of human faces or other meaningful objects with complex appearance. Judd \textit{et al.} \cite{judd2009learning} studied the human gaze patterns over thousands of images and shows that people spontaneously focus their attention on image areas with certain objects. Such observation indicates that top-down influences exist as a long-term supervision and might affect human viewing behavior with or without explicit task instructions.     

\subsubsection{Saliency Detection as a Top-down Problem}
To fully model the top-down influences, recent eye-fixation prediction methods are mostly based on supervised or semi-supervised machine learning techniques. Typically, a saliency model is obtained by training a pixel-wise eye-fixation predictor, which can  be  formulated as the following optimization problem given $I$ and $Y =Y_{\text{eye}}\cup Y_{\text{rnd}}$ with Euclidean Loss:
\vspace{-0.05cm}
\begin{equation}
\begin{split}
\text{argmin}_{S,\lambda}&  \sum_{(x,y)\in Y}\left(\psi(x,y)-\delta\left(S(x,y)-\lambda\right) \right)^2,\\ 
\delta(t)&=\left\{\begin{array}{c l} 1 & t>0 \\ 0 & \text{else} \\ \end{array} \right. , \\
\psi(p,q)&= \left\{ \begin{array}{c l} 1 & (p,q)\in Y_{\text{eye}} \\ 0 & \text{else} \\ \end{array} \right. ,\\
\end{split}
\label{eq:problemDef}
\end{equation}

where the saliency map $S$ is taken as a binary classifier, and $\lambda$ is the classification threshold. In such context, the goal is to learn a model which can generate the optimal $S$ that can best discriminate the ground-truth eye-fixation points and those random ones. For theoretical convenience, most works avoid the optimization of $\lambda$ by introducing \textit{Area Under ROC Curve} (\textbf{AUC}) as the final performance measure. 

Different from traditional binary classification problem, the distribution of the positive samples in eye-fixation prediction task is very sparse, while the distribution of the negatives is much denser. Thus, direct use of the discrete fixation points might not be the most optimal solution. 

As discussed in \cite{bruce2016deeper}, the ground truth eye-fixations could be assumed to be the observations sampled from a latent distribution. Thus, one can convert discrete fixation points $\psi$ to a probabilistic density map $\psi'$ via convolution with a 2D Gaussian kernel $G_\sigma$. Let $\psi'=\psi * G_\sigma$ be the ground-truth saliency map, the above optimization problem can be further updated as follows:  
\vspace{-0.05cm}
\begin{equation}
\begin{split}
\text{argmin}_S \sum_{x,y} \left( S(x,y)-\psi'(x,y) \right)^2 
\end{split}
\label{eq:problemDef2}
\end{equation}

Here, Mean Square Error (\textbf{MSE}) is introduced as the loss function because it has a simple form and can be well explained. However, other complex loss functions can also be adopted, e.g. Softmax, Entropy and Information Gain loss.

\subsection{Overview of the Proposed Framework}
It has been widely accepted that visual saliency can be driven by both bottom-up (unsupervised) and top-down (supervised) cues. Thus, it is important to take both parts into consideration to construct a plausible and effective model. As shown in Figure \ref{fig:Overview}, the proposed saliency framework mainly contains two independent pathways. i.e. the Semantic-Aware Saliency (\textbf{SAS}) and the Contrast-Aware Saliency (\textbf{CAS}).
The \textbf{CAS} pathway is based on on-line sparse feature learning and information maximization, which learns an adaptive representation for the input image and discovers the high contrast patterns within the visual context. The \textbf{SAS} pathway aims to capture the attractive semantic information in images, especially for the presence of meaningful objects and object parts such as human faces. The two pathways characterize both long-term (semantics in natural images) and short-term (contrast in single image) attention cues and are integrated using maxima normalization. We provide two different \textbf{SAS} implementations including an End-to-End deep neural network solution and a dynamic feature integration solution, resulting in the \textbf{SCA} and \textbf{SCAFI} model respectively. 

\subsection{Semantic-Aware Saliency based on End-to-End Network}
\label{sec:sasnet}
We first introduce a straight-forward implementation of semantic-aware saliency using a customized End-to-End deep convolutional network. This \textbf{End-to-End SAS} module is derived from \textbf{VGG} \cite{chatfield2014return}, an existing deep convnet originally trained for large-scale image classification. The \textbf{VGG} features learned from large-scale dataset such as ImageNet are highly correlated with the semantic information such as objects (deep-layer) or object parts (shallow-layer). There are many applications that are based on the pre-trained \textbf{VGG} net, covering both low-level tasks like edge prediction \cite{xie2015holistically} and high-level problems such as video event detection \cite{Li2017Hierarchical}. 

Figure \ref{fig:Overview} (\textbf{Part I}) shows the architecture of the \textbf{End-to-End} \textbf{SAS} network. We follow a similar parameter setting of \cite{pan2016shallow}, which used the first 3 weight layers of a pre-trained \textbf{VGG} net, followed by a pooling layer and 6 convolutional layers. A  deconvolution layer is adopted at the end of the network to obtain the semantic-aware saliency map that matches the size of the input. For any input image, we resize its size to [$240\times 320$] because human beings can recognize most of the important objects in images at this resolution. Besides, a larger resolution means much more parameters and will significantly increase the training and testing time. 

We used the data from \textbf{SALICON} dataset to train our \textbf{SAS} net. \textbf{SALICON} is currently the largest dataset available for saliency prediction task, which provides 10K images for training and 5K images for testing. Compared to traditional eye-fixation datasets, \textbf{SALICON} is much larger in size, which is more suitable for the training of deep convnet with large numbers of parameters. However, the fixation labels of \textbf{SALICON } are obtained from mouse-clicks instead of using eye-tracking devices, which means they are not pure ground-truth and this might produce some unknown side-effects. 

We train the \textbf{SAS} network based on 9K images from the \textbf{SALICON} training set and use the rest 1K for validation. The first 3 convolutional layers are initialized based on the pre-trained \textbf{VGG} net and the remaining layers are initialized randomly. Standard preprocessing procedures are adopted including mean-subtraction and [-1,1] normalization on both the input images and the ground-truth saliency maps. For training, we adopted stochastic gradient descent with Euclidean loss using a batch size of 2 images for 24K iterations. $L^2$ weight regularizer was used for weight decay and the learning rate was halved after every 100 iterations.   

The presented \textbf{End-to-End} \textbf{SAS} network aims to capture the semantic information in images, especially for the presence of meaningful objects and object parts. The network is very light compared to those designed for object recognition \cite{chatfield2014return} but much more complex than traditional feature-based saliency models. Some intermediate results generated based on the proposed \textbf{SAS} net are presented in Figure \ref{fig:SASCAS}.

\subsection{Semantic-Aware Saliency by Feature Integration}
Inspired by the Feature Integration theory \cite{treisman1980feature} of visual attention, we propose an alternative solution for the computation of semantic-aware saliency, which directly utilizes the feature maps from the convolutional layers of \textbf{VGG} net to generate the \textbf{SAS} saliency map. 

\begin{figure}[htb]
	\begin{center}
		\includegraphics[width=0.9\linewidth]{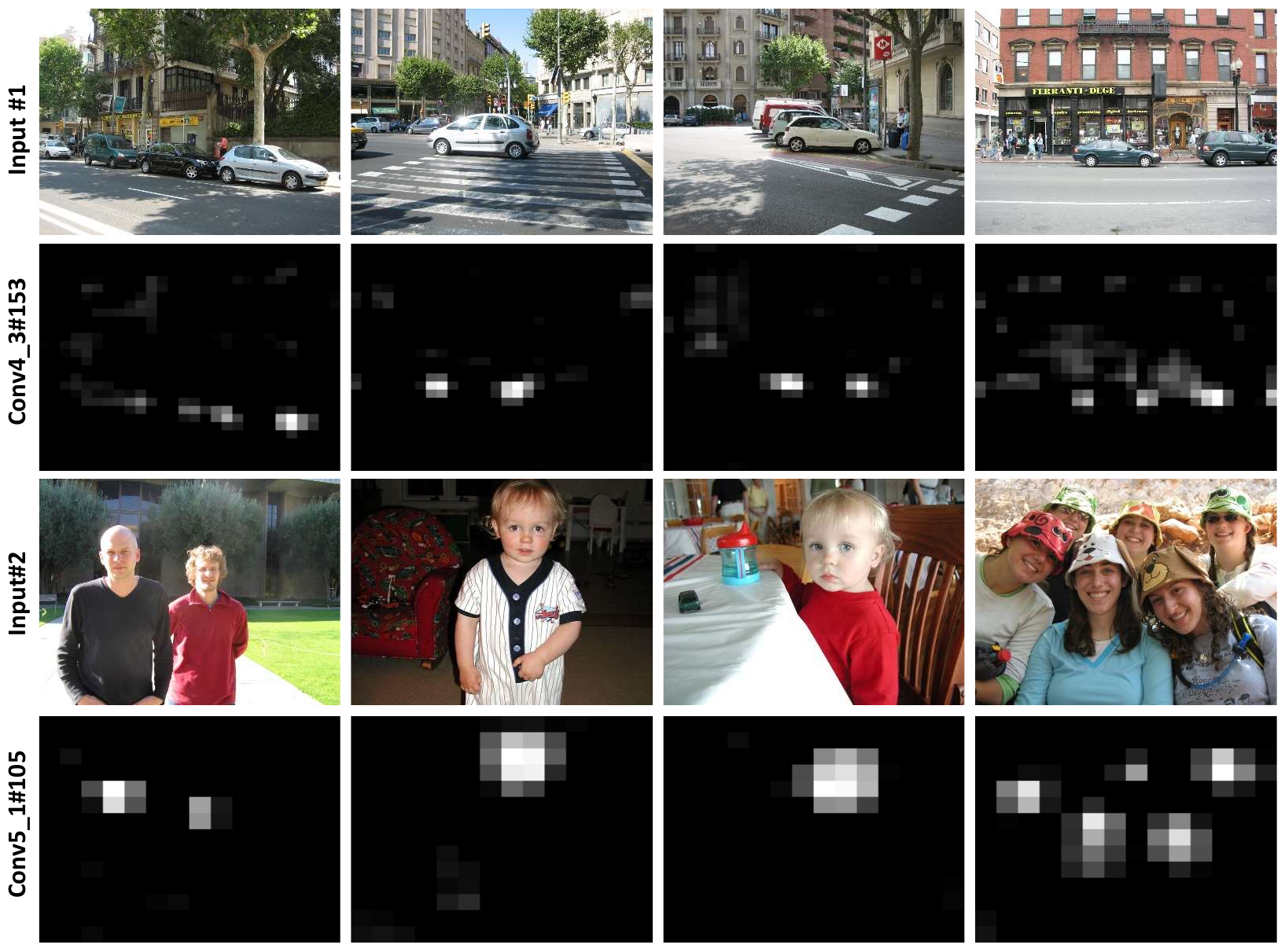}\vspace{-0.1cm}
	\end{center}
	\caption{Examples of the feature maps extracted using pre-trained \textbf{VGG} network. It clearly shows that the \textbf{VGG} features can effectively highlight important objects with certain semantics. e.g. car wheels (\#1) and faces (\#2). }
	\label{fig:Vggfea}
\end{figure}

The architecture of our dynamic feature integration model is illustrated in Figure \ref{fig:Overview} (\textbf{Part II}). After preprocessing, the RGB input image is fed into the pre-trained \textbf{VGG} network \cite{chatfield2014return}. Then we collect feature maps from all the convolutional layers except the final three fully connected layers. Each filter is corresponding to a single feature map generated based on the output of its Relu layer. All feature maps are resized to match the spatial resolution of the input. 
Figure \ref{fig:Vggfea} shows some examples of the extracted feature maps. It clearly shows that the \textbf{VGG} features can effectively highlight the semantic objects, e.g. car wheels and faces, in each input image. 

\begin{figure}[htb]
	\begin{center}
		\includegraphics[width=0.9\linewidth]{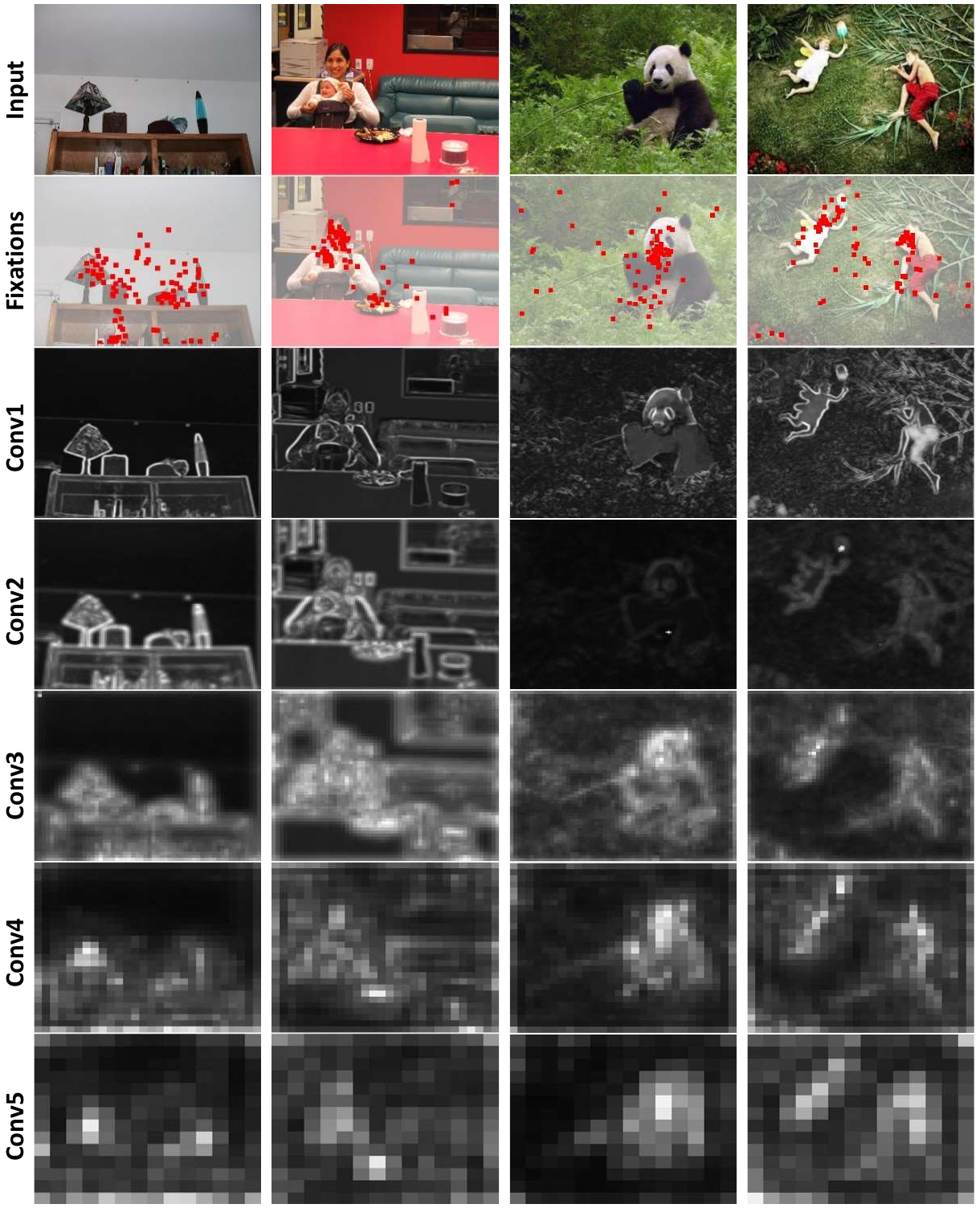}
		\vspace{-0.4cm}
	\end{center}
	\caption{Feature maps extracted from different layers of the pre-trained \textbf{VGG} network. The shallow layers contain more structural details while the deep layers highlight more about the object and object parts with specific semantics. }
	\label{fig:VggfeaScale}
\end{figure}

According to the layer configuration of the original \textbf{VGG} net, the extracted feature maps can be further classified into 5 categories, each of which corresponds to a feature set at a certain spatial scale. Table \ref{tb:vggfea} shows the key properties of each feature category, including the number of features, the original size of the feature maps and the size of the corresponding receptive field. As visualized in Figure \ref{fig:VggfeaScale}, feature maps extracted form different layers describe the image contents at different scales and in different views. The feature maps of the shallow layers contain more structural details such as edges and textures, while those from the deep layers highlight more about the object and object parts with specific semantics. 

\begin{table}[htb]
	\centering
	\caption{The Property of CNN Features from VGG-16}
	\begin{tabular}{|l|c|c|c|c|c|}
		\hline
		Layer  & \textbf{conv1} & \textbf{conv2} & \textbf{conv3} & \textbf{conv4} & \textbf{conv5} \\
		\hline
		Map Size  & 224$\times$224 & 112$\times$112 & 56$\times$56 & 28$\times$28 & 14$\times$14 \\
		\hline
		Map Number  & 128 & 256 & 768 & 1536 & 1536 \\
		\hline
		Receptive Field  & 5 & 14 & 40 & 92 & 196 \\
		\hline
	\end{tabular}
	\label{tb:vggfea}
\end{table}

We adopt a two-step procesure to generate the \textbf{SAS} saliency map based on the above \textbf{VGG} deep features. Firstly, we aggregate feature maps within each individual scale. Let $C_{lk}$ be the \textit{k}-th feature map of layer \textit{l} and $C_{lk}(x,y)$ denotes the feature response at location $(x,y)$. The \textbf{SAS} saliency map $S_l$ from layer $l$ is defined as:
   
\begin{equation}
\begin{split}
S_l(x,y) &= \sum_{k}P_{lk}(x,y), \\
P_{lk}(x,y) &= \frac{e^{M-C_{lk}(x,y)}}{\sum_{x,y} e^{M-C_{lk}(x,y)}},\\
\end{split}
\label{eq:AggFea}
\end{equation}
where $M=\text{max}_{x,y}C_{lk}(x,y)$ is the maximum of $C_{lk}$, $P_{lk}$ is the normalized map serving as a probability distribution inferred by $C_{lk}$. We may design some specific metrics to determine the weight of each individual $P$. However, we found in later experiments that using layer-wise weights is more convenient for feature aggregation since high-level semantics are mainly encoded in deeper  layers.

The second step is to aggregate the saliency information across all the scales. Specifically, the final \textbf{SAS} saliency map $S_{\text{SAS}}$ is obtained by a linear combination of the saliency maps from all layers:
\begin{equation}
\begin{split}
S_{\text{SAS}}(x,y) &= \sum_{l} w_l S_{l}(x,y).
\end{split}
\label{eq:AggLayer}
\end{equation}
The combination weight $\mathbf{w}=[w_1,w_2,...w_5]$ can be initialized using different strategies. An equal weighting strategy is suitable for general usage, where the \textbf{SAS} module is working as a single model. A spare setting where one scale is given a very high weight is more reasonable if the \textbf{SAS} saliency is serving as a sub-module of an integrated model. We show more experimental evidence in Sec.~\ref{sec:exp}.     

\subsection{Contrast-Aware Saliency}

When too much semantic information is presented in the stimuli, e.g. lots of similar cars running by, human visual attention is more likely to be attracted by the stimulus with higher feature contrast, e.g. a car with a unique color. In such scenarios, the influence of high-level semantics becomes less significant, and it's more reliable to use data-driven approaches for the estimation of saliency. This intuition leads to the second component of our saliency model, namely contrast-aware saliency (\textbf{CAS}, illustrated in Figure \ref{fig:Overview} \textbf{Part III}), which is implemented based on multi-scale adaptive sparse representation and information maximization.   

\subsubsection{Multi-Scale Adaptive Sparse Feature Extraction}
Given an image $I$, we first turn it into a multi-scale patch-based representation $\{\mathbf{X}_i |i = 1,2,3,4 \} $ by scanning $I$ with multiple sliding windows (window size $B_i\in \{1,3,5,7\}$) from the top-left to the bottom-right. $\mathbf{X}_i$ is stored as a $M_i\times N_i$ matrix, where each column vector corresponds to a reshaped RGB image patch ($M_i = B_i\times B_i\times 3$ is the size of each patch, $N_i$ is the total number of patches). For each $\mathbf{X}_i$, we apply Independent Component Analysis (\textbf{ICA}) \cite{HyvErkki1997nc}\footnote{\textbf{FastICA}: \url{https://research.ics.aalto.fi/ica/fastica/}}  to generate a complete sparse dictionary $\mathbf{W}_i = [\mathbf{w}_1^i, \mathbf{w}_2^i,...,\mathbf{w}_{M_i}^i ]$, where each basis is an independent component of $\mathbf{X}_i$. 

Given $\mathbf{w}_j^i\in \mathbf{W}_i$ as the $j$-th basis at scale $i$ , we can generate a feature vector $\mathbf{F}_j^i$ by treating  $\mathbf{w}_j^i$ as a linear filter:
\begin{equation}
\mathbf{F}_j^i= {\mathbf{w}_j^i}^{\mathbf{T}}\mathbf{X}_i
\label{eq:FeatureDecomposition}  
\end{equation}
where $\mathbf{F}_j^i(k)$ denotes the response value of $k$-th patch for the $j$-th basis at scale $i$.

\subsubsection{Saliency by Information Maximization}
Inspired by \cite{Bruce_Tsotsos06nips}, we measure the contrast-aware saliency of each patch by the self-information of its multi-scale sparse feature vector. Specifically, the \textbf{CAS} saliency value of the $t$-th patch at scale $i$ is defined as:
\begin{equation}
\begin{split}
\mathbf{S}_i(t)&=-\log \prod_{j} p_{i,j}(\mathbf{F}_j^i(t))\\
&=-\sum_{j}\log p_{i,j}(\mathbf{F}_j^i(t)),\\
\end{split}
\label{eq:SelfInfo}
\end{equation}
where $p_{i,j}(.)$ is the probability density function of the $j$-th feature at scale $i$. $p_{i,j}$ can be directly estimated using histogram method. We can obtain multiple saliency maps focusing on different salient patterns at various scales. The final \textbf{CAS} map $S_{\text{CAS}}$ is obtained by summing up all the single-scale \textbf{CAS} saliency:
\begin{equation}
\begin{split}
S_{\text{CAS}}(x,y)&= \sum_{i} S'_{i}(x,y),\\
\end{split}
\label{eq:CASMap}
\end{equation}
where $S'_{i}$ is a 2D version of $\mathbf{S}_i$. Here we omit the combination weights because the multi-scale sparse representation does not encode any semantic information and an equally weighted combination can uniformly characterize the contrasts across different scales.

\subsection{Integration of the Two Pathways} 
\begin{figure}[t]
	\begin{center}
		\includegraphics[width=0.9\linewidth]{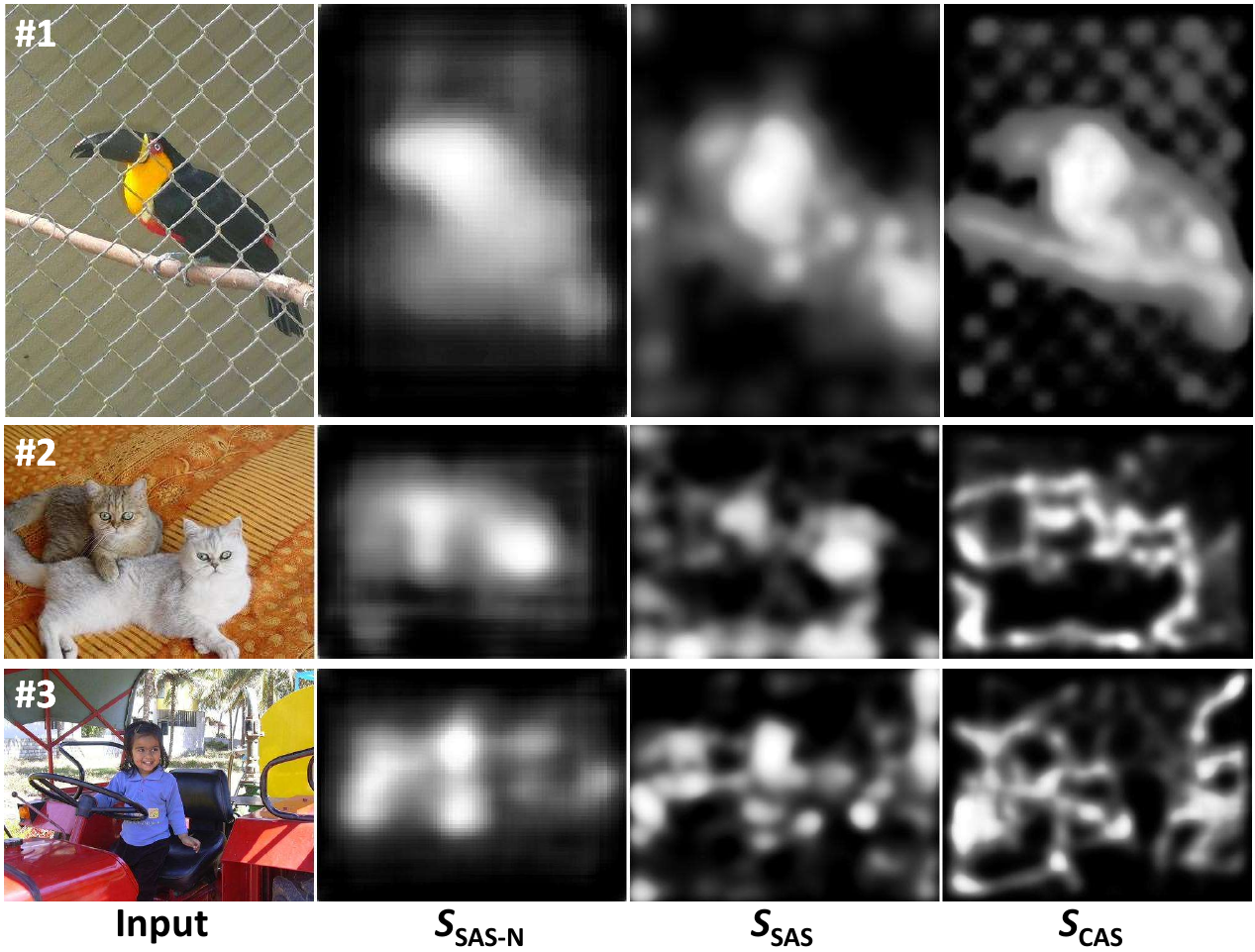}
	\end{center}
    \vspace{-0.3cm}
	\caption{Examples of $S_{\text{SAS-N}}$, $S_{\text{SAS}}$ and $S_{\text{CAS}}$ maps based on natural scenes. \textbf{SAS} maps highlight semantics while \textbf{CAS} emphasizes the contextual contrast. Image \textbf{\#1} contains one most salient object with simple background; Image \textbf{\#2} contains two salient objects; Image \textbf{\#3} consists of one human target and with complex background. $S_{\text{SAS}}$ can localize the semantic objects more accurately.}
	\vspace{-0.3cm}
	\label{fig:SASCAS}
\end{figure}

Fusing saliency detection results of multiple models has been recognized as a challenging task since the candidate models are usually developed based on different cues or assumptions. Fortunately, in our case, the integration problem is relatively easier since we only consider the outputs from two pathways. As there is no prior knowledge or other top-down guidance can be used, it's safer to utilize the map statistics to determine the importance of each pathway. Intuitively, in the final integration stage, we combine the results from two pathways by summing them up after Maxima Normalization (\textbf{MN}) (Algorithm \ref{alg:mn}):
\begin{equation}
\begin{split}
S_{\text{SCA}}= N_{max}(S_{\text{SAS-N}}) + N_{max}(S_{\text{CAS}}), \\
S_{\text{SCAFI}}= N_{max}(S_{\text{SAS}}) + N_{max}(S_{\text{CAS}}), \\
\end{split}
\label{eq:SCAFI}
\end{equation}
where $S_{\text{SAS-N}}$ is the output of the \textbf{End-to-End SAS} network (Sec. \ref{sec:sasnet}), $S_{\text{SAS}}$ and $S_{\text{CAS}}$ are generated using Eqn.~\ref{eq:AggLayer} and Eqn.~\ref{eq:CASMap} respectively. Figure \ref{fig:SASCAS} shows some visual examples of the intermediate \textbf{SAS-N}, \textbf{SAS} and \textbf{CAS} maps. It clearly illustrates the preferences and highlights of the two pathways in different image context. The \textbf{SAS} map highlights semantics such as faces and object parts, while the \textbf{CAS} map emphasizes those visual contents with high local contrast. Intuitively, \textbf{SAS} seems to be more accurate then \textbf{SAS-N} for the localization of semantic contents.

\begin{algorithm}
	\small
	\caption{Maxima Normalization $N_{max}(S,t)$}
	\label{alg:mn}
	\begin{algorithmic}[1]
		\REQUIRE  {2D intensity map $S$, thresh of local maxima $t=0.1$ }
		\ENSURE { Normalized Saliency Map $S_N$}
		\STATE Set the number of local maxima $N_M=0$
		\STATE Set the sum of the maxima $V_M=0$
		\STATE Set Global Maximum $G_M = 1$
		\STATE Normalizing the values in $S$ to a fixed range [0...$G_M$]
		\FORALL{ pixel $(x,y)$ of $S$ }
		\IF{$S(x,y)>t$}
		\STATE $R = \{S(i,j) |i=x-1,x+1, j=y-1,y+1\}$
		\IF{$S(x,y)> \text{max}(R)$}
		\STATE $V_M = V_M + S(x,y)$
		\STATE $N_M = N_M + 1$
		\ENDIF
		\ENDIF
		\ENDFOR
		\STATE $S_N = S\cdot (G_M - V_M / N_M)^2/G_M$
		\RETURN Normalized map $S_N$
	\end{algorithmic}
\end{algorithm}


The Maxima Normalization operator $N_{max}(.)$ was originally proposed for the integration of conspicuous maps from multiple feature channels \cite{Itti_etal98pami}, which has been demonstrated very effective and has a very convincing psychological explanation. In addition to the \textbf{MN} integration strategy, we also test two alternative methods in the experiment: Average-Pooling (\textbf{AP}) and Max-Pooling (\textbf{MP}) \cite{wang2016learning}.

\section{Experiments}
\label{sec:exp}
We evaluate our model in two tasks: 1) prediction of human eye fixation and 2) response to artificial images with controlled salient patterns. All experiments are based on public available datasets and the saliency maps generated by all the tested models can be downloaded from our project page\footnote{\url{https://sites.google.com/view/xssun}}.
\subsection{Datasets and Evaluation Metric}
For the eye-fixation prediction task, we used 6 public available dataset including: Bruce \cite{Bruce_Tsotsos06nips}, Cerf \cite{cerf2008predicting} , ImgSal \cite{li2013visual}, Judd \cite{judd2009learning}, PASCAL-S \cite{li2014secrets} and SALICON \cite{jiang2015salicon}. Each dataset contains a group of natural images and the corresponding eye-fixations captured from human subjects. Note that, among the 6 datasets, SALICON is the largest in size but its eye-fixation data is labeled by mouse clicks instead of using eye-tracking devices. Thus in our experiment, we use SALICON to train our End-to-End semantic-aware saliency module and apply the rest 5 dataset for performance evaluation. Table \ref{tb:Dataset} show some basic statistics of the 6 dataset. More details of the datasets can be found in \cite{li2014secrets} and \cite{jiang2015salicon}.

\begin{table}[t]
	\center
	\caption{Statistics of the 6 eye-fixation dataset.}
	\label{tb:Dataset}
	\begin{tabular}{| l|c| c| c |c|}
		\hline
		\text{DataSet} & Images & Subjects & Device & Year \\
		\hline
		\textbf{Bruce}\cite{Bruce_Tsotsos06nips} &   120 & 20 &  Eye-Tracker & 2006 \\
		\textbf{Cerf}\cite{cerf2008predicting} & 200 & 7  &  Eye-Tracker & 2008 \\
		\textbf{ImgSal}\cite{li2013visual} & 235 & 21 &  Eye-Tracker & 2013 \\
		\textbf{Judd\cite{judd2009learning}} & 1003 & 15 &  Eye-Tracker & 2009 \\
		\textbf{PASCAL-S}\cite{li2014secrets} & 850 & 8 &  Eye-Tracker & 2014 \\
		\textbf{SALICON}\cite{jiang2015salicon} & 15000 & 16 &  Mouse-Click & 2015\\
		\hline
	\end{tabular}\vspace{0.1cm}
\end{table}

Many evaluation metrics have been developed to measure the quality of saliency maps in predicting human eye-fixations. Following the procedure of the recent benchmarking surveys \cite{borji2013state,li2014secrets}, we use the shuffled Area Under ROC curve (\textbf{sAUC}) \cite{tatler2005visual} as the major evaluation metric in our experiment. The original \textbf{AUC} metric scores a saliency map according to its ability to separate the ground-truth eye fixations from random points. In \textbf{sAUC}, positive samples are taken from the eye-fixation of the test image, whereas the negative ones are sampled from other images. We use the GPU implementation of \textbf{sAUC} from \cite{li2014secrets} for the experiments. Post-processing parameters (e.g. blur kernel) are also optimized for each method to ensure a fair comparison. 

In addition to the eye-fixation datasets, we also test the models on a small group of artificial images consisting of controlled salient patterns. Most of the images are made based on solid evidence found in psychological experiments such as the preattentive features. Thus, we can further evaluate the plausibility of a model by checking whether it can produce a reasonable response to such images.
\subsection{Results on Eye-Fixation Prediction}
We have compared our proposed model with several classic saliency prediction algorithms including: \textbf{ITTI} \cite{Itti_etal98pami}, \textbf{AIM} \cite{Bruce_Tsotsos06nips}, \textbf{GBVS} \cite{harel2006graph}, \textbf{DVA} \cite{Hou08nips}, \textbf{SUN} \cite{zhang2008sun}, \textbf{SIM} \cite{Murray2011Saliency}, \textbf{QDCT} \cite{Schauerte2012Quaternion} \textbf{SIG} \cite{hou2012image}, \textbf{SGP} \cite{sun2014toward} and \textbf{AWS} \cite{leboran2017dynamic}. 
Despite the heuristic methods, we've also compared to some recent learning-based CNN models, including \textbf{SAL} \cite{huang2015salicon} (implemented using OpenSalicon \cite{christopherleethomas2016}), \textbf{DPN} \cite{pan2016shallow} and our previous \textbf{SCA} \cite{sun2017saliency} model. Untill now, \textbf{SAL} model is still one of the best according to the MIT saliency benchmark \cite{salMetrics_Bylinskii}. While \textbf{DPN} was the winner of the Saliency Detection competition in \textit{LSUN Challenge} (2015). Some details of the baseline models are presented in Table~\ref{tb:models} including their main ideas and the processing speed. Since the \textbf{DPN} and \textbf{SCA} model are partially trained on the SALICON dataset, we mainly report and discuss the evaluation results on the other 5 datasets.

\begin{table}[tb]
	\center
	\caption{Overview of the Tested Models.}
	\label{tb:models}
	\begin{tabular}{| r|l |c|c|}
		\hline
		\text{\textbf{Name}} & \textbf{Method} & \textbf{Year} & \textbf{Fps}\\
		\hline
		& \textbf{Heuristic Methods} & & \\
		\textbf{ITTI} \cite{Itti_etal98pami} & Integration of Low-level Features & 1998 & 5.1 \\
		\textbf{AIM} \cite{Bruce_Tsotsos06nips} & Information Maximization & 2006& 0.31 \\
		\textbf{GBVS} \cite{harel2006graph} & Random Walk on Graph & 2006& 3.0 \\
		\textbf{DVA} \cite{Hou08nips} & Incremental Coding Length & 2008& 33 \\
		\textbf{SUN} \cite{zhang2008sun} & Sparse Coding and Natural Statistics & 2008& 3.1 \\
		\textbf{SIM} \cite{Murray2011Saliency} & Low-level Vision Model & 2011& 0.55 \\
		\textbf{QDCT} \cite{Schauerte2012Quaternion} & Quaternion-based DCT &2012 & 110 \\
		\textbf{SIG} \cite{hou2012image} & Image Signature & 2012& 16 \\
		\textbf{SGP} \cite{sun2014toward} & Super-Gaussian Component Pursuit &2014& 4.6 \\
		\textbf{AWS} \cite{leboran2017dynamic} & Dynamic Whitening &2017& 0.73 \\
		\textbf{SCAFI} (\textbf{Ours}) & Semantic and Contrast Integration & 2018 & 2.0 \\
		\hline
		& \textbf{Learning-based Methods} & & \\
		\textbf{SAL} \cite{christopherleethomas2016}  & Open Code of SALICON\cite{huang2015salicon}  & 2015& 0.03 \\
		\textbf{DeepGazeII} \cite{kummerer2016deepgaze} & Pretrained VGG + Read-out Net & 2016& - \\
		\textbf{DPN} \cite{pan2016shallow} & Shallow and Deep Neural Network & 2016& 0.33 \\
		\textbf{SCA} \cite{sun2017saliency} & Integration of Two Saliency Cue & 2017 & 0.32 \\
		\hline
	\end{tabular}
\vspace{-0.3cm}
\end{table}

\begin{figure*}[tbh]
	\begin{center}
		\includegraphics[width=0.94\linewidth]{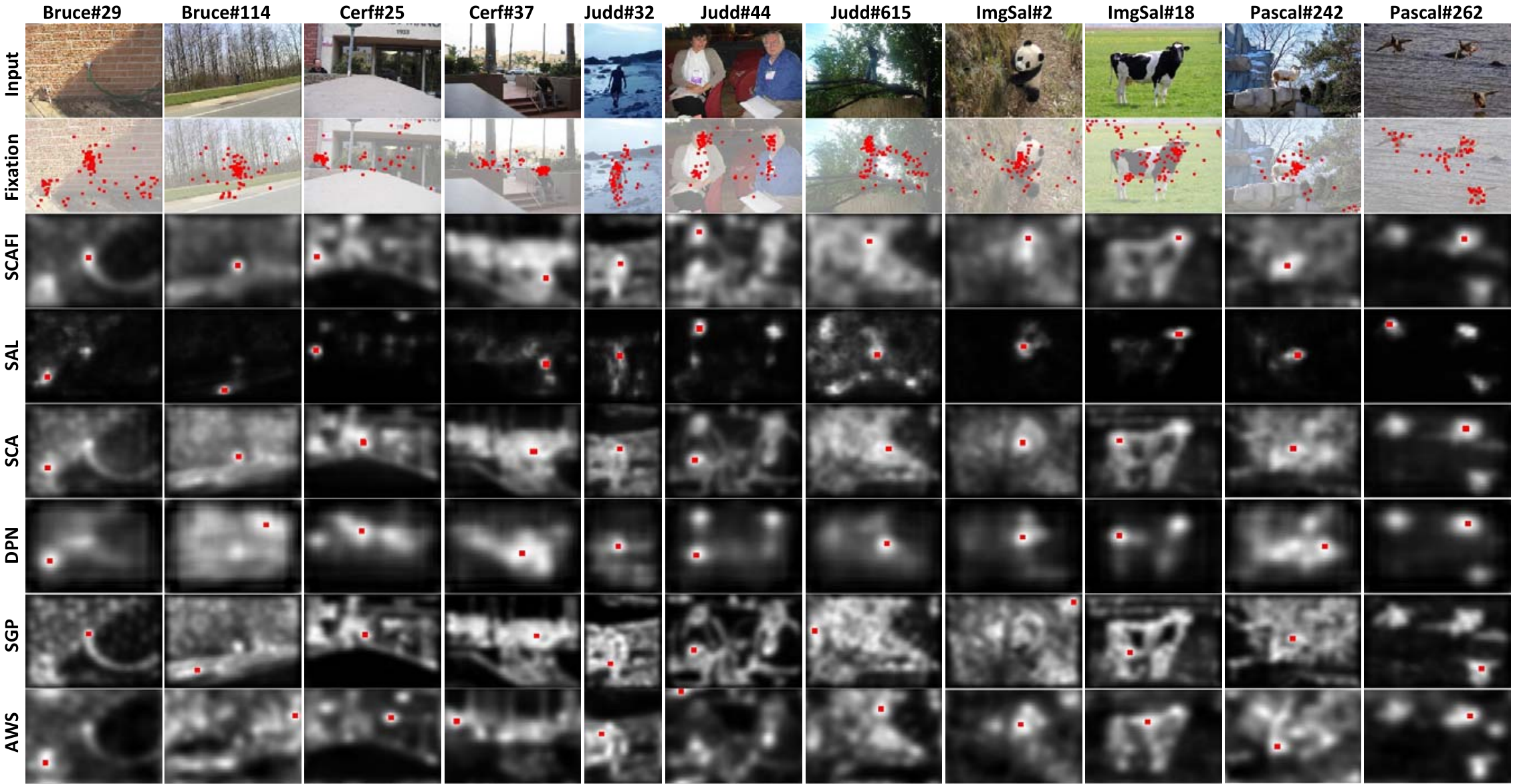}
	\end{center}
	\caption{Visual comparisons of different saliency models. The red dot in each saliency map indicates the location of the maximum saliency. The proposed  \textbf{SCAFI} model can effectively locate the most salient patterns in various conditions, while other approaches are more likely to be influenced by complex backgrounds (Judd\textbf{\#615}, Pascal\textbf{\#242})  and size variations of the salient objects (Bruce\textbf{\#114}, Cerf\textbf{\#37}, ImgSal\textbf{\#2}).} 
	\label{fig:Visual}
	\vspace{-0.3cm}
\end{figure*}

Figure~\ref{fig:Visual} shows some representative visual examples of the saliency maps generated by the tested models for various natural scenes. In each map, we use a red dot to highlight the location of the maximum saliency value. Compared to the baselines, the proposed \textbf{SCAFI} model is much more reliable and can accurately locate the most salient pattern in each image under different interference. Most traditional approaches failed when the salient object is small (Cerf\textbf{\#37}) or with a cluttered background (Judd\textbf{\#615}, Pascal\textbf{\#242}).

\subsubsection{Overall sAUC Performance} 
\begin{figure*}[htb]
	\begin{center}
		\includegraphics[width=0.9\linewidth]{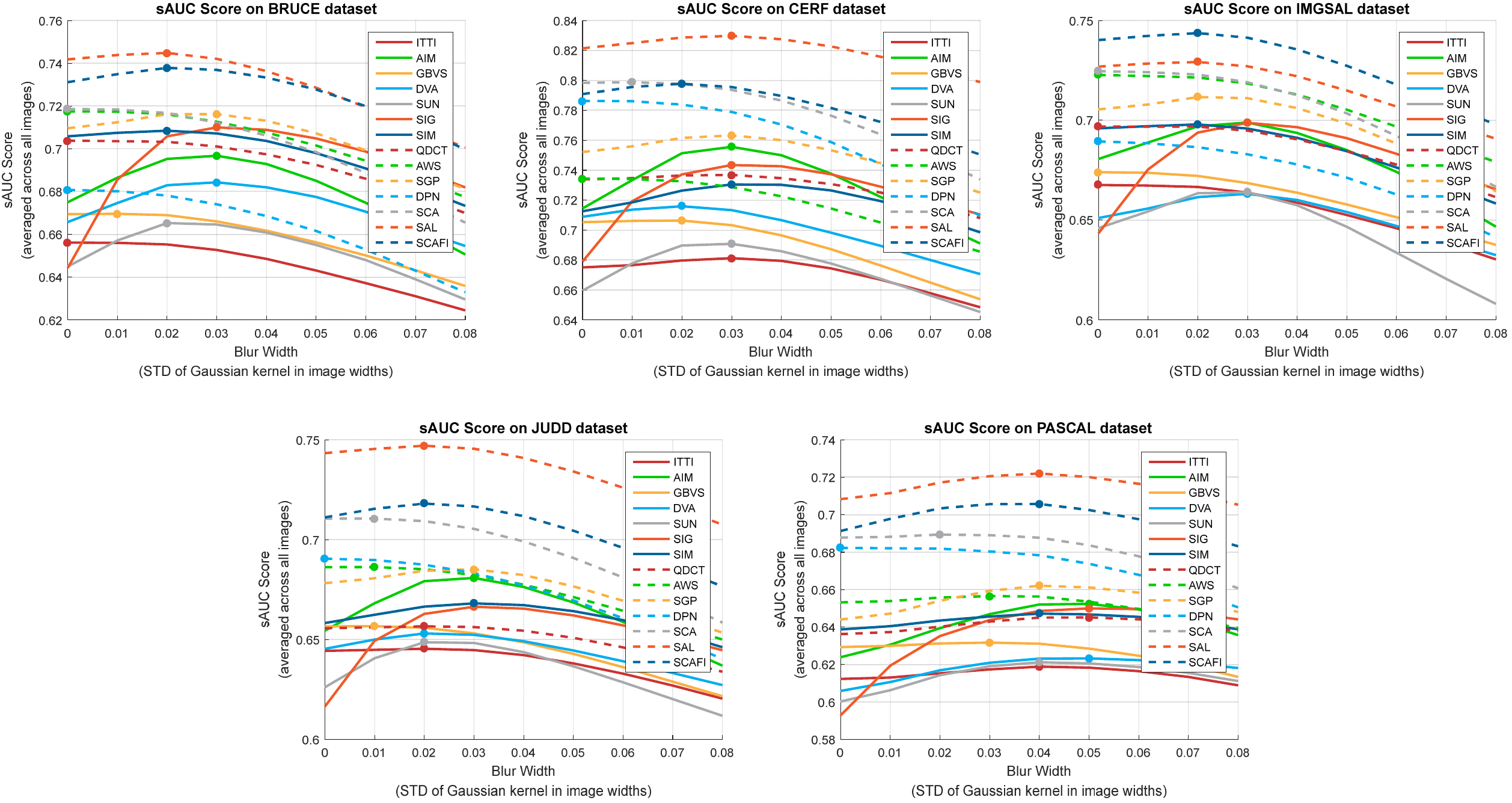}
	\end{center}
	\caption{Overall performance for eye-fixation prediction on 5 datasets. Curves depict the  \textbf{sAUC} score as a function of a blurring kernel $\sigma$. The filled dot on each curve represents the maximum \textbf{sAUC} score the corresponding model. By using deep features, \textbf{DPN}, \textbf{SCA} , \text{SAL} and \textbf{SCAFI}, significantly outperform traditional approaches on the two largest datasets, \textbf{Judd} and \textbf{PASCAL}. Besides, the impact of contrast-aware saliency is bigger in \textbf{Bruce} and \textbf{ImgSal} than in other datasets, which could explain the competitive performance of the \textbf{SGP} and \textbf{AWS} model. The proposed SCAFI is the only heuristic model that can outperform all learning-based models (on \textbf{ImgSal}).}
	\label{fig:AUCResults}
\end{figure*}

\begin{figure}[tb]
	\begin{center}
		\includegraphics[width=0.85\linewidth]{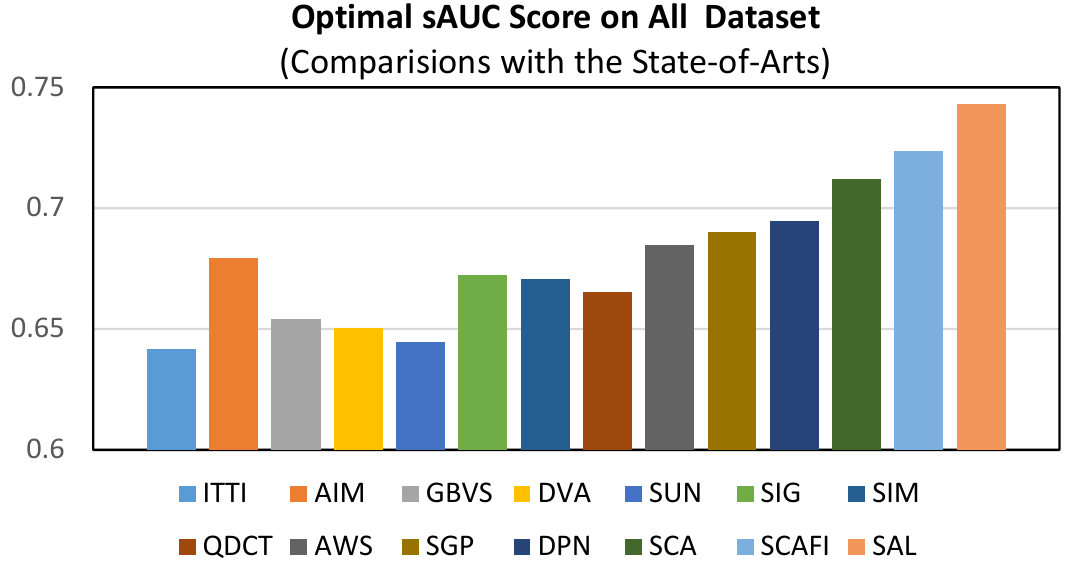}	
	\end{center}	
	\caption{The overall \textbf{sAUC} performance of the tested models on all 5 datasets. As an heuristic method, the proposed \textbf{SCAFI} gains 4.16\% improvements against \textbf{DPN}, the winner of  \textit{LSUN Challenge} (2015). With the second best sAUC performance, our \textbf{SCAFI} is 70 times faster than \textbf{SAL}.}\vspace{-0.3cm}
	\label{fig:AUCHis}
\end{figure}

Figure~\ref{fig:AUCResults} and Table~\ref{tb:AUC} presents the main \textbf{sAUC} results. We show the performance curves of the tested models over different blur kernel $\sigma$. Table \ref{tb:AUC} gives the best \textbf{sAUC} score of each model under its optimal  $\sigma$. Figure~\ref{fig:AUCHis} uses bars to show the overall \textbf{sAUC} performance of the tested models on all the datasets. Our proposed \textbf{SCAFI} achieved the best sAUC score among the heuristic models and showed competitive performance with the best learning-based model (\textbf{SAL}). 
It's also important to note that the deep models, including \textbf{DPN}, \textbf{SCA}, \textbf{SAL} and \textbf{SCAFI}, significantly outperform traditional approaches on the two largest datasets, \textbf{Judd} and \textbf{PASCAL}. This is quite reasonable because all these deep models apply pre-trained deep neural network to embed semantic cues into the saliency computation process.

\subsubsection{Influence of The Integration Strategies} Table \ref{tb:AUCFusion} shows the optimal \textbf{sAUC} score of \textbf{SCA} and \textbf{SCAFI} with Maxima Maximization (\textbf{MN}) and the other two alternative integration strategies, including \textbf{AP} (average pooling), and \textbf{MP} (max pooling). Overall, \textbf{MN} performs the best, but only slightly better than \textbf{AP}. \textbf{MN} and \textbf{AP} are much better than \textbf{MP} on all datasets. It is also worth mentioning that \textbf{MN} works more reliably for \textbf{SCAFI} compared to the previous \textbf{SCA} model.

\subsubsection{Influence of CAS and the SAS weights} 
\begin{figure*}[htb]
	\begin{center}
		\includegraphics[width=0.9\linewidth]{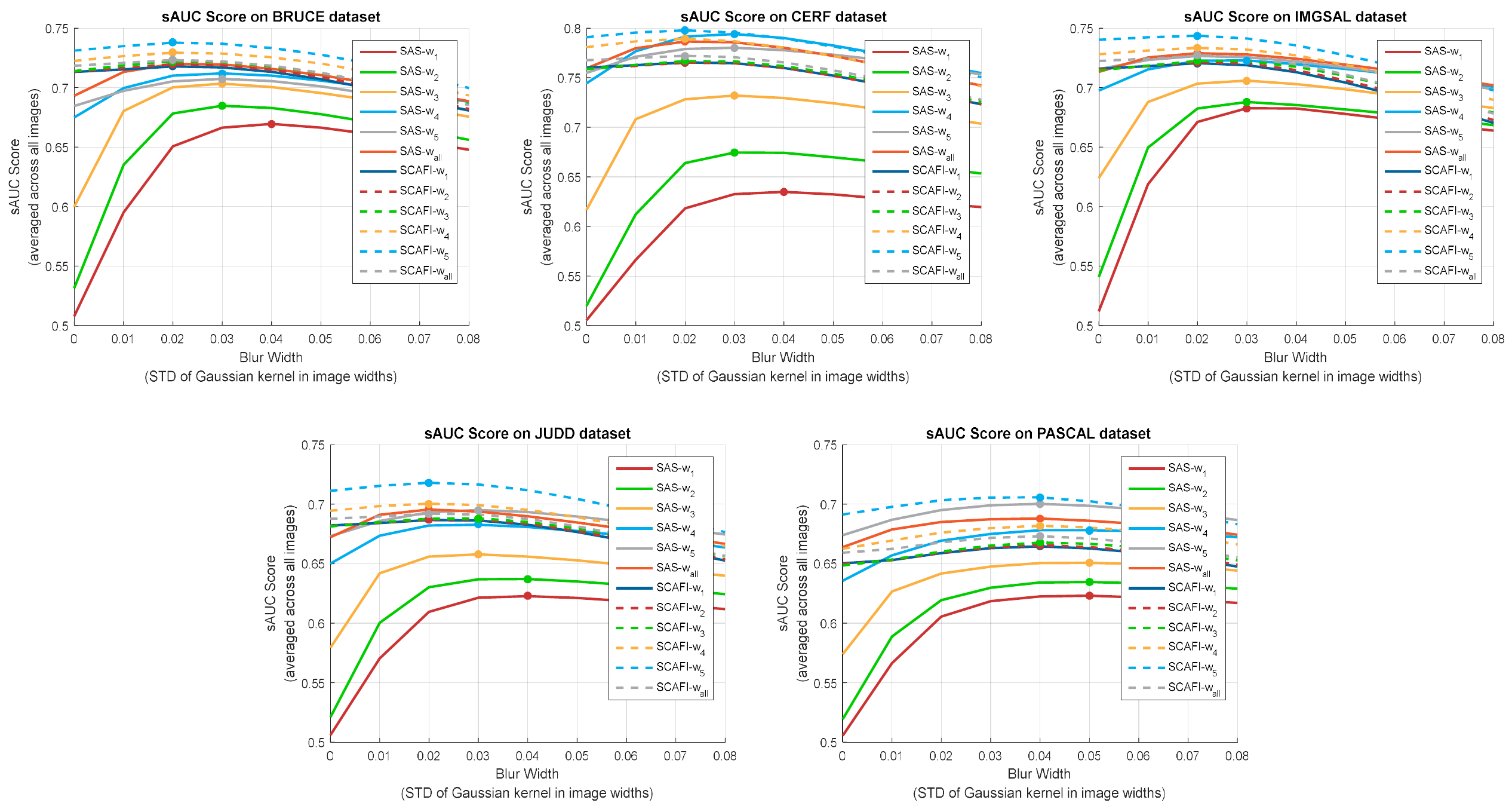}
	\end{center}
	\caption{Performance of \textbf{SCA} and \textbf{SCAFI} with different SAS weights.Curves depict the  \textbf{sAUC} score as a function of a blurring kernel $\sigma$. The filled dot on each curve represents the maximum \textbf{sAUC} score the corresponding model. \textbf{SCAFI} with $w_5$ achieves the overall best performance. }
	\label{fig:AUCSCAFI}
\end{figure*}

In Eqn.~\ref{eq:AggLayer}, we introduce a weighting parameter $\mathbf{w}$ which balances the importance of features from different \textbf{VGG} layers. We evaluate 6 intuitive configurations of $\mathbf{w}$, which corresponds to 5 groups of single-scale \textbf{VGG} features and a combination of all \textbf{VGG} features. Figure~\ref{fig:AUCSCAFI} and Table~\ref{tb:AUCWeight} shows the \textbf{sAUC} performance of \textbf{SAS} (Eqn.~\ref{eq:AggLayer} as a single model), \textbf{CAS} (Eqn.~\ref{eq:SelfInfo}) and \textbf{SCAFI} (Eqn.~\ref{eq:SCAFI}) with different \textbf{SAS} weight $\mathbf{w}$ (Table~\ref{tb:SASWeights}). For \textbf{SAS}, $\mathbf{w}_{all}$ outperforms all others on 3 datasets including \textbf{Bruce}, \textbf{ImgSal} and \textbf{Judd}, but $\mathbf{w}_5$ achieves the best overall performance. As a combination of \textbf{SAS} and \textbf{CAS}, \textbf{SCAFI} with $\mathbf{w}_5$ consistently outperforms the others on all 5 dataset. These results indicate that the \textbf{Conv5} layer of \textbf{VGG} net contains the best complementary semantic information for our \textbf{CAS} module, and their combination leads to the best accuracy for human eye-fixation prediction task.       

\subsubsection{Saliency Prediction Speed} All the tested models are implemented on a single machine with an i7-7700k 4.20GHz CPU, 32G RAM, and a Titan X Pascal GPU. As shown in Table~\ref{tb:models}, he speed of the proposed \textbf{SCAFI} is approximately 2 fps which is acceptable for most of the potential applications, e.g. image editing and enhancement. Note that, as a heuristic model, \textbf{SCAFI} can run 70$\times$ faster than \textbf{SAL} with competitive sAUC performance.

\begin{table*}[htb]
	
	\center
	\caption{The optimal \textbf{sAUC} score of the tested models on all eye-fixation datasets. }
	\scriptsize
	\begin{tabular}{|l|c c c c c c c c c c c |c c c|}
		\hline
		& \multicolumn{11}{c|}{\textbf{Heuristic Methods}}  & \multicolumn{3}{c|}{\textbf{Learn-based Methods}}\\
		& \textbf{ITTI}&\textbf{ AIM}&\textbf{ GBVS}& \textbf{DVA}& \textbf{SUN}& \textbf{SIG}& \textbf{SIM}&\textbf{ QDCT}& \textbf{AWS}& \textbf{SGP}& \textbf{SCAFI}& \textbf{DPN}& \textbf{SCA}& \textbf{SAL}\\ 
		\hline
		\textbf{Bruce} & 0.6561& 0.6967& 0.6696& 0.6843& 0.6653& 0.7100& 0.7085& 0.7038& 0.7174& 0.7162&\textbf{ 0.7378}& 0.6808& 0.7186& \textcolor{red}{\textbf{0.7448}}\\ 
		\textbf{Cerf} & 0.6813& 0.7556& 0.7063& 0.7159& 0.6908& 0.7435& 0.7305& 0.7368& 0.7344 & 0.7632& \textbf{0.7977}& 0.7863& 0.7987& \textcolor{red}{\textbf{0.8297}}\\ 
		\textbf{ImgSal} & 0.6676& 0.6988& 0.6740& 0.6632& 0.6640& 0.6987& 0.6979& 0.6969& 0.7227& 0.7116&\textcolor{red}{\textbf{ 0.7437}}& 0.6894& 0.7246& \textbf{0.7292}\\ 
		\textbf{Judd} & 0.6454& 0.6808& 0.6567& 0.6530& 0.6487& 0.6664& 0.6681& 0.6566& 0.6863& 0.6851&\textbf{ 0.7181}& 0.6905& 0.7105& \textcolor{red}{\textbf{0.7471}}\\ 
		\textbf{PASCAL} & 0.6188& 0.6524& 0.6317& 0.6233& 0.6212& 0.6500& 0.6472& 0.6452& 0.6565& 0.6621& \textbf{0.7057}& 0.6823& 0.6894& \textcolor{red}{\textbf{0.7220}}\\ 
		\hline
		\textbf{Avg.}& 0.6417& 0.6795& 0.6543& 0.6503& 0.6448& 0.6723& 0.6708& 0.6655& 0.6849& 0.6876& \textbf{0.7238}& 0.6949& 0.7122& \textcolor{red}{\textbf{0.7432}}\\ 
		\hline
	\end{tabular}\vspace{0.1cm}
	\label{tb:AUC}
\end{table*}

\begin{table}[htb]
	\center
	\caption{\textbf{SCA} and \textbf{SCAFI} with different integration strategies}
	\scriptsize
	\begin{tabular}{ |l|c c c |c c c |}
		\hline
		& \multicolumn{3}{c|}{\textbf{SCA +}}  & \multicolumn{3}{c|}{\textbf{SCAFI +}}\\
		& \textbf{AP} & \textbf{MP} & \textbf{MN} & \textbf{AP} & \textbf{MP} & \textbf{MN} \\
		\hline
		\textbf{Bruce} & \textbf{0.7194}& 0.7180& 0.7186 & 0.7371& 0.7263&\textcolor{red}{\textbf{ 0.7378}}\\ 
		\textbf{Cerf} & 0.7984& 0.7868& \textcolor{red}{\textbf{0.7987}}& 0.7977& 0.7777& \textbf{0.7977}\\ 
		\textbf{Imgsal} & \textbf{0.7257}& 0.7249& 0.7246& 0.7423& 0.7293& \textcolor{red}{\textbf{0.7437}}\\ 
		\textbf{Judd }& 0.7102& 0.7047& \textbf{0.7105}& 0.7174& 0.7000& \textcolor{red}{\textbf{0.7181}}\\ 
		\textbf{PASCAL}& 0.6879& 0.6850& \textbf{0.6894}& 0.7014& 0.6835& \textbf{0.7057}\\ 
		\hline
		\textbf{Avg.}& 0.7116& 0.7072& \textbf{0.7122}& 0.7218& 0.7048& \textcolor{red}{\textbf{0.7238}}\\ 
		\hline
	\end{tabular}
	\label{tb:AUCFusion}
\end{table}

\begin{table}[htb]
	\center
	\caption{The pre-defined SAS weights}
	\begin{tabular}{ |l|c | c | c |c  |c  |c |}
		\hline
		& $\mathbf{w}_1$ & $\mathbf{w}_2$ & $\mathbf{w}_3 $ & $\mathbf{w}_4$ & $\mathbf{w}_5$ & $\mathbf{w}_{all}$ \\ 
		\hline
		\textbf{conv1} & 1 & 0 & 0 & 0 & 0 & 0.2 \\
		\textbf{conv2} & 0  & 1 & 0 & 0 & 0 & 0.2 \\
		\textbf{conv3} & 0 & 0 & 1 & 0 & 0 & 0.2 \\
		\textbf{conv4 }& 0 & 0 & 0 & 1 & 0 & 0.2 \\
		\textbf{conv5} & 0 & 0 & 0 & 0 & 1 & 0.2 \\	
		\hline
	\end{tabular}
	\vspace{0.1cm}
	\label{tb:SASWeights}
\end{table}

\begin{table*}[htb]
	\center
	\caption{The optimal \textbf{sAUC} score of \textbf{SCAFI} based on different \textbf{SAS} weights}
	\scriptsize
	\begin{tabular}{| l|c|c c c c c c | c c c c c c | }
		\hline
		& \textbf{CAS} & \multicolumn{6}{c|}{\textbf{SAS} by Feature Integration} &\multicolumn{6}{c|}{ \textbf{SCAFI} (\textbf{SAS} + \textbf{CAS})} \\ 
		& & $\mathbf{w}_1$ & $\mathbf{w}_2$ & $\mathbf{w}_3 $ & $\mathbf{w}_4$ & $\mathbf{w}_5$ & $\mathbf{w}_{all}$ & $\mathbf{w}_1$ & $\mathbf{w}_2$ & $\mathbf{w}_3 $ & $\mathbf{w}_4$ & $\mathbf{w}_5$ & $\mathbf{w}_{all}$ \\  
		\hline
		\textbf{Bruce} & 0.7174 & 0.6694& 0.6847& 0.7034& 0.7120& 0.7074& \textbf{0.7200} & 0.7177 & 0.7195& 0.7221& 0.7294& \textbf{0.7378}& 0.7232\\ 
		\textbf{Cerf} & 0.7655 & 0.6349& 0.6746& 0.7321&\textbf{ 0.7941}& 0.7803& 0.7864 & 0.7654& 0.7652& 0.7668 & 0.7895&\textbf{ 0.7977}& 0.7721\\ 
		\textbf{Imgsal} & 0.7202 & 0.6827& 0.6879& 0.7059& 0.7228& 0.7266& \textbf{0.7290} & 0.7205 & 0.7214& 0.7224& 0.7334& \textbf{0.7437}& 0.7269\\ 
		\textbf{Judd} & 0.6867 & 0.6227 & 0.6371& 0.6578& 0.6829& 0.6951& \textbf{0.6954} & 0.6867& 0.6873& 0.6882& 0.7005& \textbf{0.7181}& 0.6920\\ 
		\textbf{PASCAL} & 0.6644 & 0.6230& 0.6347& 0.6507& 0.6782& \textbf{0.7003}& 0.6880 & 0.6646& 0.6656& 0.6678& 0.6818& \textbf{0.7057}& 0.6732\\ 
		\hline
		\textbf{Avg.}& 0.6902 & 0.6320& 0.6467& 0.6684& 0.6958& \textbf{0.7077}& 0.7049 & 0.6903& 0.6910& 0.6925& 0.7060& \textbf{0.7238}& 0.6970\\ 
		\hline
	\end{tabular}
	\label{tb:AUCWeight}
\end{table*}

\begin{figure}[htb]
	\begin{center}
		\includegraphics[width=0.85\linewidth]{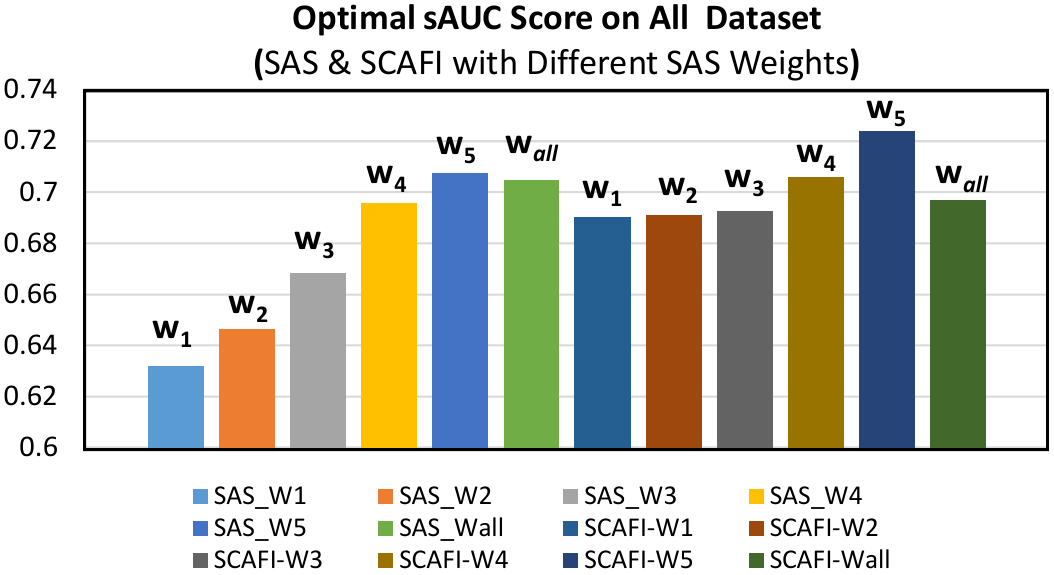}	
		\caption{The overall \textbf{sAUC} performance of \textbf{SAS} \& \textbf{SCAFI} on all 5 datasets.  The integration of adaptive sparse features and the \textbf{Conv5} layer responses from \textbf{VGG} net (\textbf{SCAFI} with $\mathbf{w}_5$) achieves the best performance.}
		\vspace{-0.4cm}
	\end{center}
	\label{fig:SCAFIHistogram}
\end{figure}


\subsection{Response to Salient Patterns}
\begin{figure*}[htb]
	\centering
		\begin{minipage}[b]{\linewidth}
		\includegraphics[width=\linewidth]{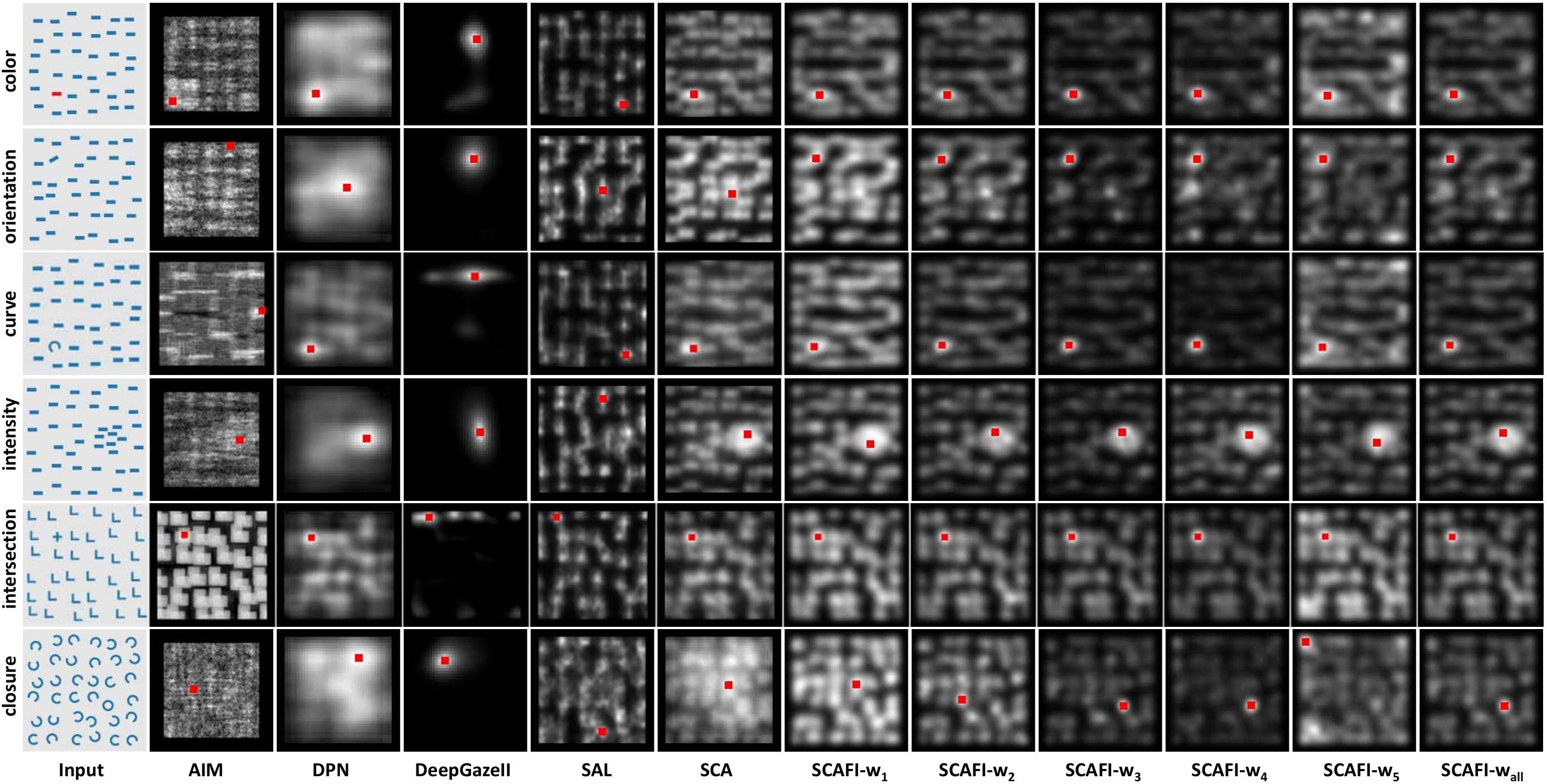}
		\centerline{\footnotesize (a) Response to artifical salient patterns}\medskip
	\end{minipage}
	\vspace{0.15cm}
	\begin{minipage}[b]{\linewidth}
		\centering
		\includegraphics[width=\linewidth]{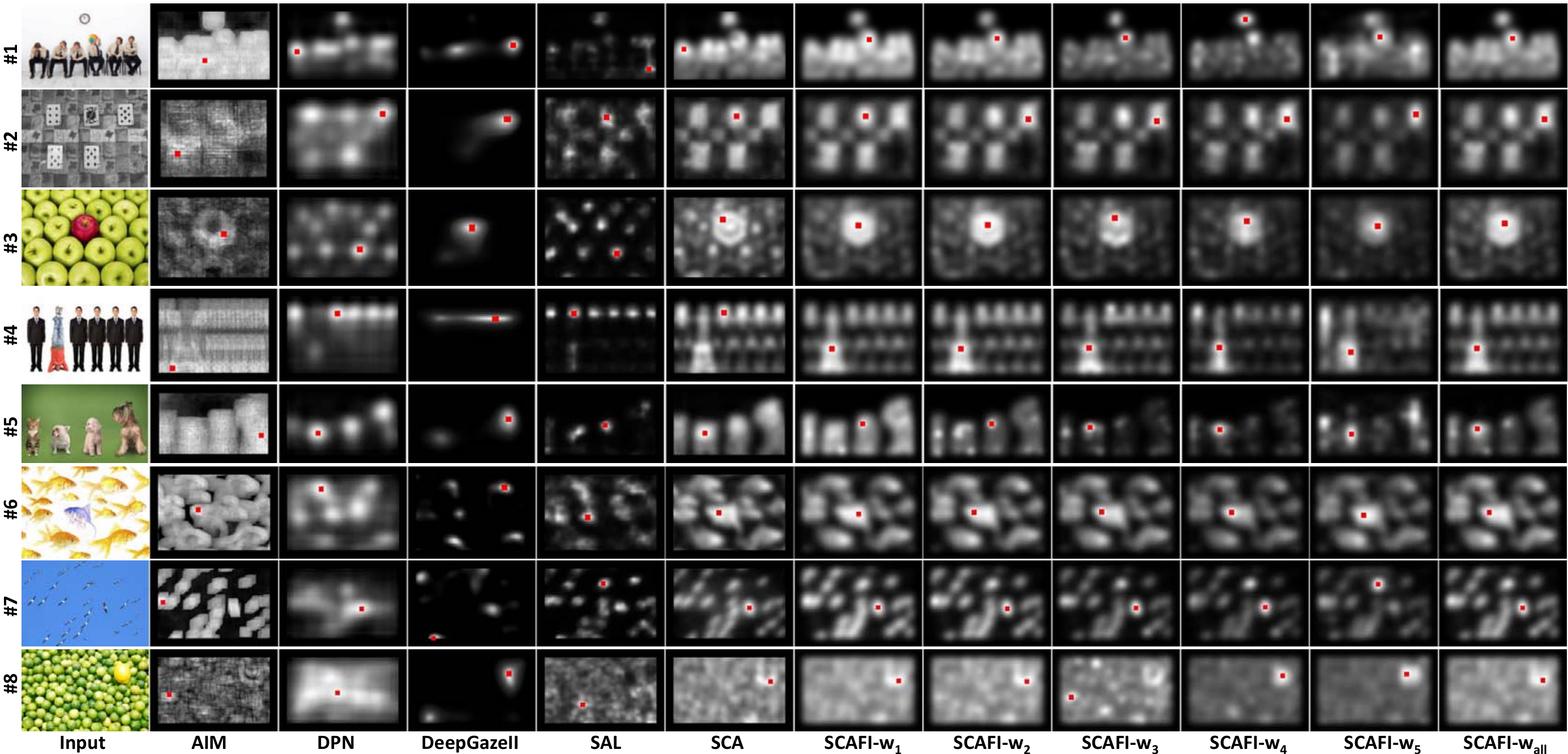}
		\centerline{\footnotesize (b) Response to natural images with controlled salient patterns}\medskip
	\end{minipage}
	\caption{Response to salient patterns. The red dot in each map indicates the location of the maximum saliency. In comparison to the results of \textbf{AIM}\cite{BruceTsotsos2009jov}, \textbf{DPN}\cite{pan2016shallow} and \textbf{DeepGazeII} \cite{kummerer2016deepgaze}, \textbf{SAL} \cite{christopherleethomas2016}, our \textbf{SCA} and \textbf{SCAFI} model successfully locate most of the salient patterns in these images, which further demonstrate their effectiveness and psychophysical plausibility. Note that, \textbf{SCAFI} with $\mathbf{w}_{1}$ and $\mathbf{w}_{2}$ are not the optimal solution for eye-fixation prediction, yet they output the most reasonable results in both (a) and (b). The results show a similar decision-making behavior of our \textbf{SCAFI} model to the actual human vision system, and thus better differentiate our method from the traditional End-to-End deep models. }
	\label{fig:Psy}
\end{figure*}

It has been well accepted that the response to the artificial patterns adopted in attention related experiments can indicate the biological plausibility of the tested saliency models. 
In Figure~\ref{fig:Psy}, we show the main results based on the images provided by Hou \textit{et al}. \cite{Hou2007saliencydetection}, Bruce \textit{et al.} \cite{bruce2016deeper} and Li \textit{et al.}\cite{li2013visual}. The red dot in each map indicates the location of the maximum saliency value. In comparison to the results of \textbf{AIM}~\cite{BruceTsotsos2009jov}, \textbf{DPN}~\cite{pan2016shallow},  \textbf{DeepGazeII}~\cite{kummerer2016deepgaze}, \textbf{SAL} \cite{christopherleethomas2016}, \textbf{SCA} and our \textbf{SCAFI} model successfully locate most of the salient patterns in the testing images, which further demonstrate their effectiveness and psychophysical plausibility. Specifically, for the basic patterns (Figure~\ref{fig:Psy}(a)), \textbf{SCAFI} highlights all the salient regions while \textbf{SCA} only fails for the case of ``closure''. Given natural images with more complex configurations (Figure~\ref{fig:Psy}(b)), both \textbf{SCAFI} and \textbf{SCA} successfully captured the most attractive objects with only a few explainable failures. Generally, \textbf{SCAFI} with  $\mathbf{w}_{1}$ and $\mathbf{w}_{2}$ output the most reasonable results, although they are not the optimal choice for the eye-fixation prediction task. 
In comparison, \textbf{DPN}, \textbf{DeepGazeII}, and \textbf{SAL} only made the most recognizable objects pop out which merely considered the object-level contrast given the image context. As we discussed in Sec.~\ref{sec:intro}, objects indeed attract human attention and draw a large proportion of eye-fixations, yet an object detector alone doesn't make a robust eye-fixation predictor because the context of the visual environment is sometimes more important for human vision system to decide the next move of the eyes.       


\section{Conclusion}
\label{sec:con}
In this paper, we proposed a heuristic framework to integrate both semantic-aware and contrast-aware saliency which combines bottom-up and top-down cues simultaneously for effective eye fixation prediction. As the middle results, the \textbf{SAS} and \textbf{CAS} maps generated by the two pathways clearly show their preference and highlights in the input image. Experimental results on 5 benchmark datasets and artificial images demonstrate the superior performance and better plausibility of the proposed \textbf{SCAFI} model over both classic approaches and the recent deep models.


\clearpage


\ifCLASSOPTIONcaptionsoff
  \newpage
\fi



\bibliographystyle{IEEEtran}
\bibliography{A1}

\end{document}